\documentclass{article}

\PassOptionsToPackage{numbers, sort&compress}{natbib}

     \usepackage[final]{neurips_2021}

\usepackage[utf8]{inputenc} 
\usepackage[T1]{fontenc}    
\usepackage[hidelinks,colorlinks = true,
            linkcolor =
            gray,
            urlcolor  = blue,
            citecolor = 
            gray,
            anchorcolor = blue]{hyperref}       
\usepackage{url}            
\usepackage{booktabs}       
\usepackage{amsfonts}       
\usepackage{nicefrac}       
\usepackage{microtype}      
\usepackage{xcolor}         
\usepackage{multirow}
\usepackage{array}
\usepackage{amsmath}
\usepackage{multirow}
\usepackage{graphicx}
\usepackage{wrapfig}
\usepackage{algorithm}
\usepackage{listings}
\usepackage{xspace}
\usepackage{eucal}

\usepackage[margin=4pt,font=small,labelfont=bf,labelsep=endash,tableposition=bottom]{caption}

\newcolumntype{H}{>{\setbox0=\hbox\bgroup}c<{\egroup}@{}}

\newlength\savedwidth

\newcommand{\blue}{\color{blue}}

\newcommand{\tabincell}[2]{\begin{tabular}{@{}#1@{}}#2\end{tabular}}
\definecolor{mygray}{gray}{.92}

\def\x{\times}
\def \eg {\textit{e.g.}\xspace}
\def \ie {\textit{i.e.}\xspace}
\def \ours          {Twins}

\def \altour   {Twins-SVT}
\def \altsmall   {Twins-SVT-S}
\def \altbase   {Twins-SVT-B}
\def \altlarge   {Twins-SVT-L}
\def \pcpvt  {Twins-PCPVT}
\def \pcpvtsmall {Twins-PCPVT-S}

\def \pcpvtbase {Twins-PCPVT-B}
\def \pcpvtlarge {Twins-PCPVT-L}

\title{Twins: Revisiting the Design of Spatial Attention in Vision Transformers}

\author{
Xiangxiang Chu$^1$, 
~~
Zhi Tian$^2$,  
~~
Yuqing Wang$^1$, 
~~ 
Bo Zhang$^1$, \\[0.1cm]
\textbf{Haibing Ren}$^1$, 
~~
\textbf{Xiaolin Wei}$^1$,
~~
\textbf{Huaxia Xia}$^1$, 
~~
\textbf{Chunhua Shen}$^2$\thanks{Corresponding author.}
\\ [0.25cm]
$^1$ Meituan Inc. ~ ~ ~ ~   $^2$ The University of Adelaide, Australia
\\[0.1cm]
{$^1$ \tt\small \{chuxiangxiang,wangyuqing06,zhangbo97,renhaibing,weixiaolin02,xiahuaxia\}@meituan.com}\\
{$^2$ \tt\small zhi.tian@outlook.com, chunhua@me.com} 
}

\begin{document}

\maketitle

\begin{abstract}
    Very 
    recently, a variety of vision transformer architectures for dense prediction tasks have been proposed and they 
    show 
    that 
    the 
    design of 
    spatial attention is 
    critical 
    to their success in these tasks. In this work, we revisit 
    the 
    design of the spatial attention and 
    demonstrate 
    that a carefully devised yet simple spatial attention
    mechanism  
    performs favorably against the state-of-the-art schemes. As a result, we propose two vision transformer architectures, namely,
    \pcpvt\ and \altour. Our proposed architectures are highly
    efficient and easy to implement, only involving 
    matrix multiplications that are highly optimized in modern deep learning frameworks. More importantly, the proposed architectures 
    achieve 
    excellent performance on a wide range of visual tasks including image-level classification as well as dense detection and segmentation. The 
    simplicity and strong 
    performance suggest that our proposed architectures 
    may 
    serve as stronger backbones for  many 
    vision 
    tasks. Our
    Code 
    is available 
    at: \href{https://github.com/Meituan-AutoML/Twins}{\texttt{https://git.io/Twins}}.

\end{abstract}

\section{Introduction}

Recently, Vision Transformers~\cite{dosovitskiy2021an, touvron2020training, carion2020end} have received increasing research interest. Compared to the widely-used convolutional neural networks (CNNs) in visual perception, Vision Transformers enjoy great flexibility in modeling 
 long-range  dependencies
in 
vision tasks, introduce less inductive bias,  and can 
naturally process 
multi-modality input 
data 
including 
images, videos, texts, speech signals, and point clouds. Thus, they have been considered to be a strong alternative to CNNs.
It is expected that 
vision transformers 
are likely to replace CNNs and 
serve as the most basic component in the next-generation visual perception systems.

One of the prominent problems when applying transformers to 
vision 
tasks is the heavy computational complexity incurred by the spatial self-attention operation in transformers, which grows quadratically 
in the 
number of pixels 
of 
the input image. 
A workaround is the \emph{locally-grouped self-attention} (or self-attention in non-overlapped windows as 
in the recent Swin Transformer \cite{liu2021swin}), where the input is spatially grouped into non-overlapped windows and the standard self-attention is computed only within each sub-window. Although it can significantly reduce the complexity, it lacks the connections between different windows and thus results in a limited receptive field. As 
pointed out 
by many previous works \cite{chen2017deeplab, peng2017large, liu2015parsenet}, a sufficiently large receptive field is crucial to the performance, particularly 
for 
dense prediction tasks such as image segmentation and object detection. Swin
\cite{liu2021swin}
proposes a shifted window operation to 
tackle 
the issue, where the boundaries of these local windows are gradually moved as the network proceeds.  Despite being effective, the shifted windows may have uneven sizes.  The uneven windows result in difficulties when the models are deployed with ONNX or TensorRT, which prefers the windows 
of 
equal sizes. Another solution is proposed in PVT~\cite{wang2021pyramid}. Unlike the standard self-attention operation, where each query computes the attention weights with all the input tokens, in PVT, each query only computes the attention with a sub-sampled version of the input tokens. Although its computational complexity in theory is still quadratic, it is already 
manageable  
in practice.

From a unified perspective, the core in the aforementioned vision transformers is {how the spatial attention is designed}.  Thus, in this work, we revisit the design of the spatial attention  in vision transformers. 
Our first 
finding 
is that the global sub-sampled attention in PVT is highly effective, and with the applicable positional encodings~\cite{chu2021ConditionalPE}, its performance can be {on par} or even better than 
state-of-the-art vision transformers (\eg, Swin). This results in our first proposed architecture, termed {\it \pcpvt}.
On top of that, we further propose a \emph{carefully-designed yet simple spatial attention} mechanism, making our architectures more efficient than PVT. Our attention mechanism is inspired by the widely-used separable depthwise convolutions and thus we name it \emph{spatially separable self-attention} (SSSA). Our proposed SSSA is 
composed of two 
types 
of attention operations---(i) \emph{locally-grouped self-attention} (LSA), and (ii) \emph{global   sub-sampled   attention} (GSA), where LSA captures the fine-grained and short-distance information and GSA deals with the long-distance and global 
information. This leads to the 
second
proposed
vision transformer architecture, 
termed 
{\it \altour}. It is worth noting that both attention operations in the architecture are \emph{efficient and easy-to-implement} with matrix multiplications in a few lines of code. Thus, all of our architectures here have great applicability and can be easily deployed.

We benchmark our proposed architectures on a number of visual tasks, ranging from image-level classification to pixel-level semantic/instance segmentation and object detection. Extensive experiments show that both of our proposed 
architectures 
perform favorably against other state-of-the-art vision transformers with similar or even reduced 
computational complexity.

\section{Related Work}
\textbf{Convolutional neural networks.} Characterized by local connectivity, weight sharing, shift-invariance and pooling, CNNs have been the \textit{de facto} standard 
model for computer vision tasks.
The top-performing models \cite{%
he2016deep,tan2019efficientnet,chollet2017xception,xie2017aggregated%
%
} in image classification also serve as the strong backbones for downstream detection and segmentation tasks. 

\textbf{Vision Transformers.} Transformer was firstly proposed by \cite{vaswani2017attention} for machine translation tasks, and since then they have become the state-of-the-art models for NLP tasks, overtaking the sequence-to-sequence approach built on LSTM. Its core component is multi-head self-attention which models the relationship between input tokens and shows great flexibility.

In 2020, 
Transformer 
was
introduced to 
computer vision 
for image and 
video processing
\cite{dosovitskiy2021an,touvron2020training,chu2021ConditionalPE,han2021transformer,ramachandran2019stand,xu2021coscale, wang2020end,
chen2020pre,
yang2020learning,zeng2020learning,dai2020up,Srinivas2021BottleneckTF,carion2020end,zheng2020rethinking,parmar2018image,touvron2021going,yuan2021tokens,jiang2021token,srinivas2021bottleneck,chen2021crossvit,wu2021cvt,xu2021coscale,zhu2021deformable}. 
 In the image classification task, 
  ViT \cite{dosovitskiy2021an} and DeiT \cite{touvron2020training}
 divide the images into patch embedding sequences and feed them into the standard transformers.
Although vision transformers have been proved compelling in image classification compared with CNNs, 
a
challenge 
remains when it is applied to dense prediction tasks such as object detection and segmentation. These tasks 
often 
require feature pyramids for better processing objects of different scales, and take as inputs the high-resolution images, which significantly increase the computational complexity of the self-attention operations.
%
%

Recently, Pyramid Vision Transformer (PVT) \cite{wang2021pyramid} is proposed and can output the feature pyramid \cite{lin2017feature} as in CNNs. PVT has demonstrated 
good 
performance in a number of dense prediction tasks. 
The recent 
Swin Transformer \cite{liu2021swin} introduces non-overlapping window partitions and restricts self-attention within each local window, resulting in linear computational complexity in the number of input tokens. To interchange information among different local areas, its window partitions are particularly 
designed 
to shift between two adjacent self-attention layers. The semantic segmentation framework OCNet~\cite{OCNet2021} shares some similarities with us and they also interleave the local and global attention. Here, we demonstrate this is a general design paradigm in vision transformer backbones rather than merely an incremental module in semantic segmentation.


\textbf{Grouped and Separable Convolutions.} Grouped convolutions are originally proposed in AlexNet \cite{krizhevsky2012imagenet} for distributed computing.
They were 
proved both efficient and effective in speeding up the networks. As an extreme case, depthwise convolutions \cite{chollet2017xception,sifre2014rigid} use the number of groups that is equal to the input or output channels, which is followed by point-wise convolutions to aggregate the information 
across 
different channels. Here, the proposed spatially separable self-attention shares some similarities with them. 

\textbf{Positional Encodings.}
Most vision transformers use absolute/relative positional encodings, 
depending on 
downstream tasks, which are based on sinusoidal functions \cite{vaswani2017attention} or learnable \cite{dosovitskiy2021an,touvron2020training}. In CPVT~\cite{chu2021ConditionalPE}, the authors propose the conditional positional encodings, which are dynamically conditioned on the inputs and show better performance than the absolute and relative ones.

\section{Our Method: \ours}
We present two simple yet powerful spatial designs for vision transformers. The
first method 
is built  
upon 
PVT \cite{wang2021pyramid} and CPVT \cite{chu2021ConditionalPE}, which only uses the global attention. The architecture is 
thus termed 
\pcpvt. 
The second 
one, termed 
\altour, is based on the proposed SSSA which interleaves local and global attention.

\subsection{\pcpvt}
PVT \cite{wang2021pyramid} introduces the pyramid multi-stage design to better 
tackle 
dense prediction 
tasks such as object detection  and semantic segmentation. It inherits the absolute positional encoding  designed in ViT \cite{dosovitskiy2021an} and DeiT \cite{touvron2020training}. All layers utilize the global attention mechanism and rely on  spatial reduction to cut down the computation cost of 
processing 
the whole sequence. It is surprising to see that the recently-proposed Swin 
transformer
\cite{liu2021swin}, which is based on shifted local windows, can perform considerably better than PVT, even on 
dense prediction tasks where a 
sufficiently 
large 
receptive field is even more 
crucial to good performance. 

In this work, we surprisingly found that the less favored performance of PVT is 
mainly 
due to the \textit{absolute positional encodings} 
employed in PVT  \cite{wang2021pyramid}. 
%
As shown in CPVT~\cite{chu2021ConditionalPE}, the absolute positional encoding 
encounter  
difficulties in 
processing 
the inputs with varying sizes (which are common in dense prediction tasks). Moreover, this positional encoding also breaks the \emph{translation invariance}. On the contrary, Swin transformer makes use of the relative positional encodings, which bypasses the above issues. Here, we 
demonstrate 
that this is the main cause why Swin outperforms PVT, and we 
show that if the 
appropriate 
positional encodings are used, PVT can actually achieve on par or even better performance than the Swin transformer.

Here, we use the conditional position encoding (CPE) proposed in CPVT \cite{chu2021ConditionalPE} to replace the absolute 
PE 
in PVT. CPE is conditioned on the inputs and can naturally avoid the above issues of the absolute encodings.  The position encoding generator (PEG) \cite{chu2021ConditionalPE}, 
which generates the CPE, is placed after the first encoder block of each stage. We use the simplest form of PEG, \ie, a 2D depth-wise convolution without batch normalization. For image-level classification, following CPVT, we remove the class token and use global average pooling (GAP) at the end of the stage \cite{chu2021ConditionalPE}. For other 
vision 
tasks, we follow the design of PVT. \pcpvt\ inherits the advantages 
of 
both PVT and CPVT, which makes it easy to be implemented efficiently. 
Our 
extensive experimental results show 
that 
this simple design can match the performance of the recent state-of-the-art Swin
transformer.  We have also attempted to replace the relative PE with CPE in Swin, which however does not result in noticeable performance gains, as shown in our experiments. We conjecture that this 
maybe 
due to the use of shifted windows in Swin, which might not work well with CPE.

\paragraph{Architecture settings} We report the detailed settings of \pcpvt\ 
in Table~\ref{tab: \pcpvtbase} (in supplementary), which are similar to PVT \cite{wang2021pyramid}. Therefore, \text{\pcpvt} has similar FLOPs and number of parameters to \cite{wang2021pyramid}.

\begin{figure}
   \includegraphics[width=0.98\linewidth]{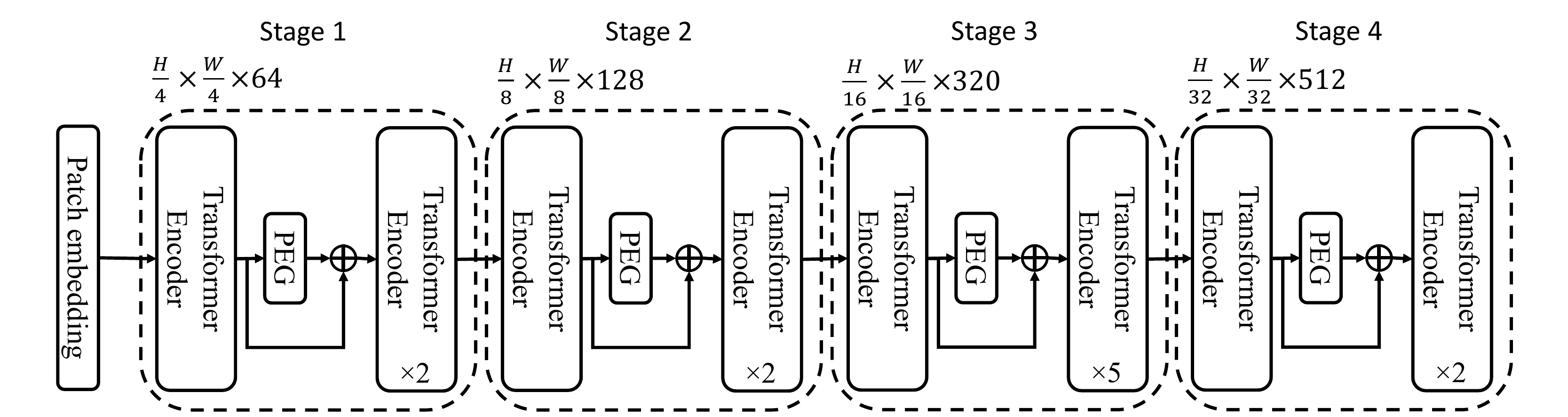}
   \caption{Architecture of Twins-PCPVT-S. ``PEG" is the positional encoding generator from CPVT~\cite{chu2021ConditionalPE}.}
   \label{fig:twins-pcpvt-s-arch}
\end{figure}

\subsection{\altour}
Vision transformers suffer severely from the heavy computational complexity in dense prediction tasks due to high-resolution inputs. Given an input of $H \times W$ resolution,  the complexity of self-attention with dimension $d$ is $\mathcal{O}(H^2W^2d)$. Here, we propose the spatially separable self-attention (SSSA) to 
alleviate 
this challenge. SSSA is composed of locally-grouped self-attention (LSA) and global   sub-sampled   attention (GSA).

\begin{wrapfigure}{R}{0.5\columnwidth}
\centering
\includegraphics[width=0.9\linewidth]{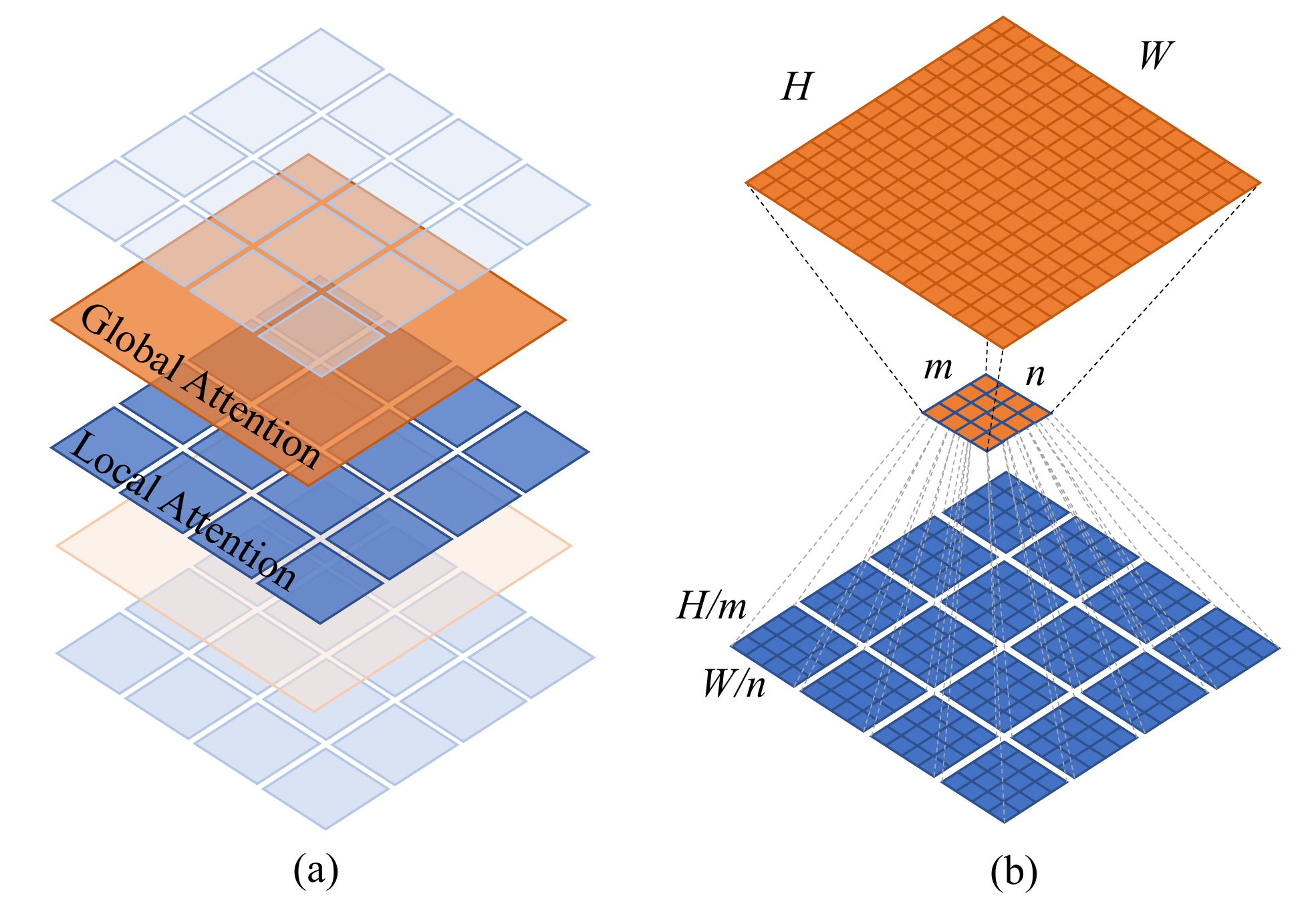}
\caption{\textbf{(a)} \altour {} interleaves locally-grouped attention (LSA) and global   sub-sampled   attention (GSA). \textbf{(b)} Schematic view of the locally-grouped attention (LSA) and global  sub-sampled   attention (GSA).}
\label{fig:local-global-attention}
\vskip -0.2in
\end{wrapfigure}

\paragraph{Locally-grouped self-attention (LSA).} 
Motivated by the group design in depthwise convolutions for efficient inference, we first equally divide the 2D feature maps into sub-windows, making self-attention communications only happen within each sub-window. This design also resonates with the multi-head design in self-attention, where the communications only occur within the channels of the same head. To be specific, the feature maps are divided into $m\times n$ sub-windows. Without loss of generality, we 
assume 
$H \% m = 0$ and $W \% n = 0$. Each group contains $\frac{HW}{mn}$ elements, and thus the computation cost of the self-attention in this window is $\mathcal{O} (\frac{H^2W^2}{m^2n^2}d)$, and the total cost is $\mathcal{O} (\frac{H^2W^2}{mn}d)$. If we let $k_1=\frac{H}{m}$ and $k_2 = \frac{W}{n}$, the cost can be 
computed 
as $\mathcal{O} (k_1k_2HWd)$, which is significantly more 
efficient 
when 
$k_1 \ll H$ and $k_2 \ll W$ and grows linearly with $HW$ if $k_1$ and $k_2$ are fixed.

Although the locally-grouped self-attention mechanism is computation friendly, the image is divided into non-overlapping sub-windows. Thus, we need a mechanism to communicate between different sub-windows,  as in Swin. Otherwise, the information  
would 
be limited to be processed locally, which makes the receptive field 
small and significantly degrades the performance as shown in our experiments. This resembles the fact that we cannot 
replace all standard convolutions by 
depth-wise convolutions in  CNNs. 

\paragraph{Global   sub-sampled   attention (GSA).} A simple solution is to add extra standard global self-attention layers after each local attention block,
which can enable cross-group information exchange. 
However, this approach would come with the computation complexity of $\mathcal{O}(H^2W^2d)$.

Here, we use a single representative to summarize the important information for each of $m \x n$ sub-windows and the representative is used to communicate with other sub-windows (serving as the key in self-attention), which can dramatically reduce the cost to
$\mathcal{O} (mnHWd)=\mathcal{O} (\frac{H^2W^2d}{k_1k_2})$. This is essentially equivalent to using the   sub-sampled   feature maps as the key in attention operations, and thus we 
term 
it global   sub-sampled   attention (GSA). If we alternatively use the aforementioned LSA and GSA like separable convolutions (depth-wise + point-wise). The total computation cost is $\mathcal{O} (\frac{H^2W^2d}{k_1k_2}+k_1k_2HWd)$.
We have 
$\frac{H^2W^2d}{k_1k_2}+k_1k_2HWd \geq 2HWd\sqrt{HW}$. The minimum is obtained when $k_1 \cdot k_2=\sqrt{HW}$. We note that $H=W=224$ is popular in classification. Without loss of generality, we use square sub-windows, \textit{i.e.},  $k_1=k_2$. Therefore, $k_1=k_2=15$ is 
close 
to the global minimum for $H=W=224$. However, our network is designed to include several stages with variable resolutions. Stage 1  has feature maps of 56 $\times$ 56, the minimum is obtained when $k_1=k_2=\sqrt{56}\approx7$.  Theoretically, we can calibrate optimal $k_1$ and $k_2$ for each of the stages. 
For simplicity, 
we use $k_1=k_2=7$ everywhere.  As for stages with lower resolutions, we control the summarizing  window-size of GSA to avoid too small amount of generated keys. Specifically, we use the size of 4, 2 and 1 for the last three stages respectively. 


As for the sub-sampling function, we investigate several options including average pooling, depth-wise strided convolutions, and regular strided convolutions. Empirical results show that regular strided convolutions perform best here. 
Formally, our spatially separable self-attention (SSSA) can be written as
\begin{equation}
	\begin{split}
	&{{\hat{\bf{z}}}^{l}_{ij}} = \text{LSA}\left( {\text{LayerNorm}\left( {{{\bf{z}}^{l - 1}_{ij}}} \right)} \right) + {\bf{z}}^{l - 1}_{ij}, \\
&{{\bf{z}}^l_{ij}} = \text{FFN}\left( {\text{LayerNorm}\left( {{{\hat{\bf{z}}}^{l}}_{ij}} \right)} \right) + {{\hat{\bf{z}}}^{l}_{ij}},\\
&{{\hat{\bf{z}}}^{l+1}} = \text{GSA}\left( {\text{LayerNorm}\left( {{{\bf{z}}^{l}}} \right)} \right) + {\bf{z}}^{l}, \\
&{{\bf{z}}^{l+1}} = \text{FFN}\left( {\text{LayerNorm}\left( {{{\hat{\bf{z}}}^{l+1}}} \right)} \right) + {{\hat{\bf{z}}}^{l+1}}, \\
& i \in \{1, 2, ...., m\}, j \in \{1, 2, ...., n\}\\
\label{eq:gvt}
	\end{split}
\end{equation}
where LSA means locally-grouped self-attention within a sub-window;
GSA is the global   sub-sampled   attention by interacting with the representative keys (generated by the sub-sampling functions) from each sub-window $\hat{\bf{z}}_{ij} \in \mathcal{R}^{k_1\times k_2 \times C}$. Both LSA and GSA have multiple heads as in the standard self-attention.The PyTorch code of LSA is given in Algorithm \ref{alg:local_group_attention} (in supplementary).



Again, we use the PEG of CPVT \cite{chu2021ConditionalPE} to encode position information and 
process 
variable-length inputs on the fly. It is inserted after the first block in each stage.

\noindent\textbf{Model variants.}
The detailed configure of \altour\ is shown in Table~\ref{tab: \altour} (in supplementary). We try our best to use the similar settings as in Swin \cite{liu2021swin} to make sure 
that 
the good  performance  is due to the new design paradigm. 

\noindent\textbf{Comparison with PVT.} PVT entirely utilizes global attentions as DeiT does while our method  makes use of  spatial separable-like design with LSA and GSA, which is more efficient.

\noindent\textbf{Comparison with  Swin.} Swin utilizes the alternation of local window based attention where the window partitions in successive layers are shifted. This is used to introduce communication among different patches and to increase the receptive field. 
However, this procedure is relatively complicated and may not 
be optimized for speed on 
devices such as mobile devices. Swin Transformer depends on torch.roll() to perform cyclic shift and its reverse on features. This operation is memory unfriendly and rarely supported by popular inference frameworks such as NVIDIA TensorRT, Google Tensorflow-Lite, and Snapdragon Neural Processing Engine SDK (SNPE), etc. This hinders the deployment of Swin either on the server-side or on end devices in a production environment. In contrast, Twins models don't require such an operation and only involve matrix multiplications that are already optimized well in modern deep learning frameworks. Therefore, it can further benefit from the optimization in a production environment. For example, we converted Twins-SVT-S from PyTorch to TensorRT , and its throughput is boosted by 1.7$\times$.  
Moreover, our local-global design 
can better exploit 
the global context, which is known to play an important role in many 
vision 
tasks.

Finally, one may note that the network configures (\eg, such as depths, hidden dimensions, number of heads, and the expansion ratio of MLP) of our two variants are sightly different. This is intended because we want to make fair comparisons to the two recent well-known transformers PVT and Swin. PVT prefers a slimmer and deeper design while Swin is wider and shallower. This difference makes PVT have slower training than Swin. \text{\pcpvt} is designed to compare with PVT and shows that a proper positional encoding design can greatly boost the performance and make it on par with recent state-of-the-art models like Swin. On the other hand, \text{\altour} demonstrates the potential of a new paradigm as to spatially separable self-attention is highly competitive to recent transformers.

\section{Experiments}
\subsection{Classification on ImageNet-1K}
We first present the ImageNet classification results with our proposed models. We carefully control the experiment settings to make fair comparisons 
against 
recent works \cite{touvron2020training,wang2021pyramid,chu2021ConditionalPE}. All our models are trained for 300 epochs with a batch size of 1024  using the  AdamW optimizer \cite{loshchilov2018decoupled}. The learning rate is initialized 
to be 
0.001 and decayed to zero within 300 epochs following the cosine strategy. We use a linear warm-up in the first five epochs and the same regularization setting as in  \cite{touvron2020training}. Note that we do not utilize extra tricks in \cite{touvron2021going,jiang2021token} to make fair comparisons although it 
may 
further improve the performance of our method. We use increasing stochastic depth \cite{huang2016deep} augmentation   of 0.2, 0.3,  0.5 for small, base and large model respectively. 
Following Swin \cite{liu2021swin}, we use gradient clipping with a max norm of 5.0 to stabilize the training process, which  is especially  important for the training of  large models.

We report the classification results on ImageNet-1K \cite{russakovsky2015imagenet} in Table~\ref{tab: classfication}. \text{\pcpvtsmall} outperforms PVT-small by $1.4\%$ and obtains similar result as Swin-T with 18\% fewer FLOPs. \text{\altsmall} is better than Swin-T with about $35\%$ fewer FLOPs. Other models 
demonstrate 
similar advantages. 

It is interesting to see that, without bells and whistles, \text{\pcpvt} 
performs 
\textit{on par} with the recent state-of-the-art Swin,  which is based on 
much more sophisticated 
designs as mentioned 
above. 
Moreover, \text{\altour} also achieves similar or better results, 
compared to 
Swin, 
indicating that the spatial separable-like design is an 
effective 
and promising paradigm. 

One may challenge our improvements are due to the use of the better positional encoding PEG. Thus, we also replace the relative PE in Swin-T with PEG \cite{chu2021ConditionalPE}, but the Swin-T's performance cannot be improved (being 81.2\%). 
\begin{table}[t]
	\caption{Comparisons with state-of-the-art  methods for ImageNet-1K classification. Throughput is tested on the batch size of 192 on a single V100 GPU. All models are trained and evaluated on 224$\times$224 resolution on ImageNet-1K dataset. $^\dagger$: w/ CPVT's position encodings  \cite{chu2021ConditionalPE}. }
	\label{tab: classfication}
	\centering
	\small 
	\begin{tabular}{ r |c|c|c|Hl}
		\toprule
		Method & Param (M) & FLOPs (G) & Throughput (Images$/$s) & Step (s)& Top-1 (\%)  \\
		\midrule
				\multicolumn{6}{c}{ConvNet} \\
				\midrule
		RegNetY-4G \cite{radosavovic2020designing} & 21& 4.0&1157 & &80.0 \\
		RegNetY-8G \cite{radosavovic2020designing}& 39 & 8.0&592&&81.7\\ 
		RegNetY-16G \cite{radosavovic2020designing}&84&16.0 &335&&82.9\\
		\midrule
		\multicolumn{6}{c}{Transformer} \\
		\midrule
		DeiT-Small/16~\cite{touvron2020training}  & 22.1 & 4.6 & 437&& 79.9 \\
		CrossViT-S ~\cite{chen2021crossvit} & 26.7 & 5.6&-&&81.0\\
		T2T-ViT-14 ~\cite{yuan2021tokens} &22& 5.2&-&&81.5\\
		TNT-S ~\cite{han2021transformer} & 23.8& 5.2&-&&81.3\\
		CoaT Mini ~\cite{xu2021coscale}&10&6.8& -&&80.8\\
		CoaT-Lite Small ~\cite{xu2021coscale} & 20 & 4.0&-&&81.9\\
		PVT-Small \cite{wang2021pyramid}  & 24.5 & 3.8 & 820& 0.57 & 79.8 \\ 
		CPVT-Small-GAP \cite{chu2021ConditionalPE} & 23 & 4.6& 817& & 81.5 \\
		\pcpvtsmall\  (\textbf{ours})& 24.1&3.8 &815& 0.57&81.2 \blue(+1.3)\\
		Swin-T \cite{liu2021swin} & 29 & 4.5& 766 & 0.48 &81.3 \\
		Swin-T + CPVT$^\dagger$ & 28&4.4&766&&81.2 \\
		\altsmall\ (\textbf{ours})& 24 &2.9& 1059 &0.47&81.7 \blue(+1.8) \\

		\midrule
		T2T-ViT-19 \cite{yuan2021tokens} & 39.2& 8.9 &-&&81.9\\
		PVT-Medium \cite{wang2021pyramid} & 44.2 & 6.7 &  526&0.80 & 81.2\\
		\pcpvtbase (ours) &43.8&6.7 &525& 0.80 &82.7 \blue(+0.8)\\
		Swin-S~\cite{liu2021swin} & 50 & 8.7 & 444& 0.73 &83.0\\
		\altbase\  (\textbf{ours})&56 &8.6& 469 & 0.73&83.2 \blue(+1.3)\\
		\midrule
		ViT-Base/16~\cite{dosovitskiy2021an} & 86.6 & 17.6 & 86& & 77.9 \\
		DeiT-Base/16~\cite{touvron2020training} & 86.6 & 17.6 & 292& & 81.8 \\
		T2T-ViT-24 \cite{yuan2021tokens} & 64.1&14.1&-&&82.3\\
		CrossViT-B ~\cite{chen2021crossvit} & 104.7 & 21.2&-&&82.2\\
		TNT-B \cite{han2021transformer} & 66& 14.1 &-&&82.8\\
		CPVT-B~\cite{chu2021ConditionalPE} & 88 & 17.6 & 292 & &82.3\\
		PVT-Large \cite{wang2021pyramid}& 61.4 & 9.8 & 367 & 1.12& 81.7 \\
		\pcpvtlarge (\textbf{ours}) &60.9 & 9.8 & 367 & &83.1 \blue(+5.2)\\
	    Swin-B \cite{liu2021swin} & 88 & 15.4& 275& 0.89 &83.3 \\
		\altlarge\  (\textbf{ours}) & 99.2& 15.1 & 288 & 0.90 & 83.7 \blue(+5.8)\\

		\midrule
		\multicolumn{6}{c}{Hybrid} \\
		\midrule
		BoTNet-S1-59 \cite{srinivas2021bottleneck} &33.5 & 7.3&-&&81.7 \\
		BossNet-T1 \cite{li2021bossnas}&- & 7.9 &-&&81.9 \\
		CvT-13 \cite{wu2021cvt} & 20 & 4.5 &-&&81.6\\
		BoTNet-S1-110 \cite{srinivas2021bottleneck} & 54.7 & 10.9&-&&82.8\\
		CvT-21 \cite{wu2021cvt}& 32 & 7.1 &-&&82.5\\

		\bottomrule

	\end{tabular}

\end{table}

\subsection{Semantic Segmentation on ADE20K}
We further evaluate the performance on segmentation tasks. We 
test on the
ADE20K dataset \cite{zhou2017scene},  a challenging scene parsing task for semantic segmentation, which is popularly 
evaluated by 
recent Transformer-based methods. This dataset contains  20K images for training and 2K images for validation. Following the common practices, we use the training set to train our models and report the mIoU on the validation set.  All  models are pretrained on the ImageNet-1k dataset.

\noindent\textbf{\pcpvt\ vs. PVT.} We compare our  \pcpvt\ with PVT~\cite{wang2021pyramid} because they have similar design and computational complexity. To make fair comparisons, we use the Semantic FPN framework \cite{kirillov2019panoptic} and exactly the same training settings as in PVT. Specifically, we train 80K steps with a batch size of 16 using AdamW \cite{loshchilov2018decoupled}. The learning rate is initialized as 1$\times$10$^{-4}$ and scheduled by the `poly' strategy with the power coefficient of 0.9. We 
apply the drop-path regularization of 0.2 for the backbone and weight decay 0.0005 for the whole network. Note that we use a stronger drop-path regularization of 0.4 for the large model to avoid over-fitting. For Swin, we use their official code  and trained models. We report the results in Table~\ref{tab: segmentation}.
With 
comparable FLOPs, \text{\pcpvtsmall} outperforms PVT-Small with a large margin (+4.5\%  mIoU), which also surpasses ResNet-50 by 7.6\%  mIoU. It also outperforms Swin-T with a clear margin. Besides, \text{\pcpvtbase} also achieves 3.3\%  higher mIoU than PVT-Medium, and \text{\pcpvtlarge} surpasses PVT-Large with 4.3\% higher mIoU.

\noindent\textbf{\altour\ vs. Swin.} We also compare our \altour\ with the recent state-of-the-art model Swin~\cite{liu2021swin}. With the Semantic FPN framework and the above settings, \text{\altsmall} achieves better performance (+1.7\%) than Swin-T. \text{\altbase} obtains comparable performance with Swin-S and \text{\altlarge}  outperforms Swin-B by 0.7\% mIoU (left columns in Table~\ref{tab: segmentation}). In addition, Swin evaluates its performance using the UperNet framework \cite{xiao2018unified}. We transfer our method to this framework and use exactly the same training settings as \cite{liu2021swin}. To be specific, we use the AdamW optimizer to train all models for 160k iterations with a global batch size of 16. The initial learning rate is 6$\times$10$^{-5}$ and linearly decayed to zero. We also utilize warm-up during the first 1500 iterations. Moreover, we apply the drop-path regularization of 0.2 for the backbone and weight decay 0.01 for the whole network.  We report the mIoU of both single scale and multi-scale testing (we use scales from 0.5 to 1.75 with step 0.25) in the right columns of Table~\ref{tab: segmentation}. 
Both with multi-scale testing, \text{\altsmall}  outperforms Swin-T by 1.3\% mIoU. Moreover, \text{\altlarge} achieves new state of the art result 50.2\%  mIoU under comparable FLOPs and outperforms Swin-B by 0.5\% mIoU. \text{\pcpvt} also achieves comparable  performance to Swin \cite{liu2021swin}.


\begin{table}
\vskip -0.25in
 \setlength\tabcolsep{5pt}
	\caption{Performance comparisons with different backbones on ADE20K validation dataset. FLOPs are tested on 512$\times$512 resolution. All backbones are pretrained on ImageNet-1k except SETR \cite{SETR}, which is pretrained on ImageNet-21k dataset.}
	\label{tab: segmentation}
	\centering
	\small 
	\begin{tabular}{r*{2}{c}l|*{3}{c}}
		\toprule
	    \multirow{2}{*}{Backbone}     & \multicolumn{3}{c|}{Semantic FPN 80k (PVT \cite{wang2021pyramid} setting)}   & \multicolumn{3}{c}{Upernet 160k (Swin \cite{liu2021swin} setting)}  \\
	    & FLOPs &Param & mIoU  & FLOPs & Param & mIoU/MS mIoU \\
	    & (G) & (M) & (\%) & (G) & (M) & (\%) \\
		\midrule
		ResNet50 \cite{he2016deep} &45 & 28.5 & 36.7 &-&-&-  \\
		PVT-Small \cite{wang2021pyramid}  &40& 28.2& 39.8  &-&-&-    \\
		\pcpvtsmall\  (ours) &40&28.4 & 44.3 \blue(+7.6)& 234&54.6& 46.2/47.5\\
		Swin-T \cite{liu2021swin} &46&31.9 &41.5&237& 59.9&44.5/45.8\\

		\altsmall\  (ours) & 37 & 28.3 & 43.2 \blue(+6.5)& 228 & 54.4& 46.2/47.1\\
		\midrule
		ResNet101 \cite{he2016deep} &66& 47.5 & 38.8& 258 &86& -/44.9\\
		PVT-Medium \cite{wang2021pyramid} &55&48.0 &41.6&-&-&- \\
		\pcpvtbase\ (ours)  & 55 &48.1&44.9 \blue(+6.1)& 250& 74.3& 47.1/48.4\\
		Swin-S \cite{liu2021swin} &70& 53.2& 45.2 &261 &81.3&47.6/49.5\\

		\altbase\ (ours) & 67& 60.4&45.3 \blue(+6.5)& 261 & 88.5 & 47.7/48.9\\
		\midrule
		ResNetXt101-64$\times$4d \cite{xie2017aggregated}& - &86.4& 40.2&-&-&-  \\
		PVT-Large \cite{wang2021pyramid}& 71 &65.1 & 42.1&-&-&- \\
		\pcpvtlarge\ (ours) & 71 &65.3& 46.4  \blue(+6.2)&269 &91.5&48.6/49.8 \\
		Swin-B \cite{liu2021swin} &107&91.2&46.0&299& 121& 48.1/49.7 \\
		\altlarge\  (ours) & 102 & 103.7& 46.7 \blue(+6.5)& 297 &133&48.8/50.2\\
		\midrule
		  Backbone   & \multicolumn{3}{c|}{PUP (SETR \cite{SETR} setting)}   & \multicolumn{3}{c}{MLA (SETR \cite{SETR} setting)} \\
		 \midrule
		T-Large (SETR) \cite{SETR}& - & 310 & 50.1 & - & 308 & 48.6/50.3\\
		
		\bottomrule
	\end{tabular}
\end{table}
\subsection{Object Detection and Segmentation on COCO}
We evaluate the performance of our method using two representative frameworks: RetinaNet \cite{lin2017focal} and Mask RCNN \cite{he2017mask}. Specifically, we use our transformer models to build the backbones of these detectors. All the models are trained under the same setting as  in \cite{wang2021pyramid}. Since PVT and Swin report their results using different frameworks, we try to make fair comparison and build consistent settings for future methods. Specifically, we report standard 1$\times$-schedule (12 epochs) detection results on the COCO 2017 dataset \cite{lin2014microsoft} in Tables~\ref{tab: detection-retina} and \ref{tab: detection-mask}. As for the evaluation based on RetinaNet, we train all the models using  AdamW \cite{loshchilov2018decoupled} optimizer for 12 epochs with a batch size of 16. The initial learning rate is 1$\times$10$^{-4}$, started with 500-iteration warmup and decayed by 10$\times$ at the 8th and 11th epoch, respectively. We use stochastic drop path regularization of 0.2 and weight decay 0.0001.  The implementation is based on MMDetection \cite{mmdetection}. For the Mask R-CNN framework,  we use the initial learning rate of 2$\times$10$^{-4}$ as in \cite{wang2021pyramid}. All other hyper-parameters follow the default settings in MMDetection. As for 3$\x$ experiments, we follow the common multi-scale training  in \cite{carion2020end,liu2021swin}, \ie, randomly resizing the input image so that its shorter side is between 480 and 800 while keeping longer one less than 1333. Moreover, for 3$\x$ training of Mask R-CNN, we use an initial learning rate of 0.0001 and weight decay of 0.05 for the whole network as \cite{liu2021swin}.

For 1$\times$ schedule object detection with RetinaNet, $\text{\pcpvtsmall}$ surpasses PVT-Small with 2.6\%  mAP and \text{\pcpvtbase} exceeds PVT-Medium by 2.4\%  mAP on the COCO \texttt{val2017} split. \text{\altsmall} outperforms Swin-T with 1.5\%  mAP while using 12\% fewer FLOPs. Our method outperform the others with similar advantage in 3$\times$ experiments. 

For  1$\times$ object segmentation with the Mask R-CNN framework, \text{\pcpvtsmall}\ brings similar improvements (+2.5\% mAP) over PVT-Small.  Compared with PVT-Medium, \text{\pcpvtbase} obtains 2.6\%  higher mAP, which is also on par with that of Swin. Both \altsmall\ and \altbase\ achieve better or slightly better performance compared to  the counterparts of Swin.
As for large models, our results are shown in Table~\ref{tab: detection-l-1x} (in supplementary) and we also achieve better performance with comparable FLOPs.

\begin{table}[ht]
 \setlength\tabcolsep{0.2pt}
		\caption{Object detection performance on the COCO \texttt{val2017} split using the RetinaNet framework. 
		1$\x$ is 12 epochs and 3$\x$ is 36 epochs. ``MS'': Multi-scale training.
		FLOPs are evaluated on 800$\times$600 resolution.} 
	\label{tab: detection-retina}
	\centering
	\small 
\begin{tabular}{rcc|lccccc|l*{5}{c}}
	\toprule
	\multirow{2}{*}{Backbone} &\multirow{2}{*}{\tabincell{c}{FLOPs\\(G)}}& \multirow{2}{*}{\tabincell{c}{Param \\(M)}} &\multicolumn{6}{c|}{RetinaNet 1$\times$} &\multicolumn{6}{c}{RetinaNet 3$\times$ + MS} \\
	\cmidrule{4-15} 
	& &&AP &AP$_{50}$ &AP$_{75}$ &AP$_S$ &AP$_M$ &AP$_L$ &AP &AP$_{50}$ &AP$_{75}$ &AP$_S$ &AP$_M$ &AP$_L$ \\
	\midrule
	ResNet50~\cite{he2016deep}&111 &37.7 & 36.3 & 55.3 & 38.6 & 19.3 & 40.0 & 48.8 & 39.0 & 58.4 & 41.8 & 22.4 & 42.8 & 51.6  \\
	PVT-Small~\cite{wang2021pyramid}& 118 & {34.2} & {40.4} & {61.3} & {43.0} & {25.0} & {42.9} & {55.7} & {42.2} & {62.7} & {45.0} & {26.2} & {45.2} & {57.2 }\\
		\pcpvtsmall\  (ours)& 118 & 34.4 &43.0\blue(+6.7)&64.1&46.0&27.5&46.3&57.3& 45.2\blue(+6.2)&66.5&48.6&30.0&48.8&58.9\\
	Swin-T \cite{liu2021swin} &118 &38.5&41.5 & 62.1& 44.2& 25.1& 44.9& 55.5& 43.9&64.8& 47.1&28.4&47.2&57.8\\
	\altsmall\  (ours)& 104 & 34.3& 43.0\blue(+6.7)& 64.2&46.3& 28.0& 46.4&57.5 & 45.6\blue(+6.6)& 67.1&48.6&29.8&49.3&60.0\\
	\midrule
	ResNet101~\cite{he2016deep} &149&56.7  & 38.5 & 57.8 & 41.2 & 21.4 & 42.6 & 51.1 & 40.9 & 60.1 & 44.0 & 23.7 & 45.0 & 53.8
	\\
	ResNeXt101-32$\times$4d~\cite{xie2017aggregated}& 151 &56.4& 39.9& 59.6 & 42.7 & 22.3 & 44.2 & 52.5 & 41.4 & 61.0 & 44.3 &23.9& 45.5 & 53.7 \\
	PVT-Medium~\cite{wang2021pyramid}  & 151 &{53.9} & {41.9}& {63.1} & {44.3} & {25.0} & {44.9} & {57.6} & {43.2} & {63.8} & {46.1} & {27.3} & {46.3} & {58.9}\\
		\pcpvtbase\  (ours)&151&54.1&44.3\blue(+5.8)&65.6 & 47.3& 27.9&47.9&59.6&46.4\blue(+5.5)&67.7&49.8&31.3&50.2&61.4 \\
		Swin-S \cite{liu2021swin} & 162 &59.8&44.5& 65.7&47.5&27.4&48.0&59.9& 46.3 &67.4& 49.8& 31.1&50.3&60.9\\

	\altbase\  (ours)& 163 & 67.0& 45.3\blue(+6.8)& 66.7&48.1&28.5&48.9&60.6& 46.9\blue(+6.0)& 68.0&50.2& 31.7& 50.3& 61.8\\
	
%
%
	\bottomrule	
\end{tabular}
\end{table}

\begin{table}
     \setlength\tabcolsep{0.1pt}
	\caption{Object detection and instance segmentation performance on the COCO \texttt{val2017} dataset using the Mask R-CNN framework. FLOPs are evaluated on a 800$\times$600 image.}
	\label{tab: detection-mask}
	\centering
	\small 
\begin{tabular}{rcc|lcclcc|lcc*{5}{c}}
	\toprule
	\multirow{2}{*}{Backbone} & \multirow{2}{*}{\tabincell{c}{FLOPs \\(G)}} & \multirow{2}{*}{\tabincell{c}{Param \\(M)}} &\multicolumn{6}{c|}{Mask R-CNN 1$\times$} &\multicolumn{6}{c}{Mask R-CNN 3$\times$ + MS} \\
	\cmidrule{4-15} 
	& &&AP$^{\rm b}$ &AP$_{50}^{\rm b}$ &AP$_{75}^{\rm b}$ &AP$^{\rm m}$ &AP$_{50}^{\rm m}$ &AP$_{75}^{\rm m}$ &AP$^{\rm b}$ &AP$_{50}^{\rm b}$ &AP$_{75}^{\rm b}$ &AP$^{\rm m}$ &AP$_{50}^{\rm m}$ &AP$_{75}^{\rm m}$ \\
	\midrule
	ResNet50~\cite{he2016deep}& 174 & 44.2& 38.0 & 58.6 & 41.4 & 34.4 & 55.1 & 36.7 & 41.0 & 61.7 & 44.9 & 37.1 & 58.4 & 40.1\\
	PVT-Small ~\cite{wang2021pyramid} &178&{44.1} & {40.4}& {62.9} & {43.8} & {37.8} & {60.1} & {40.3} & {43.0}& {65.3} & {46.9} & {39.9} & {62.5} & {42.8}\\
		\pcpvtsmall\  (ours) & 178&44.3&42.9\textsubscript{\blue(+4.9)} & 65.8 &47.1& 40.0\textsubscript{\blue(+5.6)}& 62.7& 42.9 &46.8\textsubscript{\blue(+5.8)}&69.3& 51.8 & 42.6& 66.3&46.0\\
	Swin-T \cite{liu2021swin} & 177 & 47.8& 42.2& 64.6& 46.2& 39.1&61.6& 42.0& 46.0&68.2&50.2&41.6&65.1&44.8\\

	\altsmall\  (ours)& 164 & 44.0& 43.4\textsubscript{\blue(+5.4)} & 66.0& 47.3&40.3\textsubscript{\blue(+5.9)} &63.2&43.4&46.8\textsubscript{\blue(+5.8)}&69.2& 51.2 &42.6 & 66.3 & 45.8 \\
	\midrule
	ResNet101~\cite{he2016deep} & 210 & 63.2 & 40.4 & 61.1 & 44.2 & 36.4 & 57.7 & 38.8 & 42.8 & 63.2 & 47.1 & 38.5& 60.1& 41.3\\
	ResNeXt101-32$\times$4d~\cite{xie2017aggregated}& 212&{62.8} & 41.9& 62.5 & {45.9} & 37.5 & 59.4 & 40.2 & 44.0& 64.4 & 48.0 & 39.2 & 61.4 & 41.9 \\
	PVT-Medium \cite{wang2021pyramid} &211 &63.9 & {42.0} &{64.4} &45.6 &{39.0}& {61.6}& {42.1} & {44.2}& {66.0} & {48.2} & {40.5} & {63.1} & {43.5}\\
	\pcpvtbase\  (ours)& 211 & 64.0& 44.6\textsubscript{\blue(+4.2)}& 66.7 & 48.9& 40.9\textsubscript{\blue(+4.5)}&63.8& 44.2 & 47.9\textsubscript{\blue(+5.1)} & 70.1& 52.5& 43.2 & 67.2& 46.3 \\

	Swin-S \cite{liu2021swin} & 222 & 69.1 & 44.8 & 66.6& 48.9 & 40.9 & 63.4& 44.2 & 47.6&69.4& 52.5&42.8&66.5&46.4 \\
	\altbase\  (ours)& 224 & 76.3& 45.2\textsubscript{\blue(+4.8)}&67.6& 49.3&41.5\textsubscript{\blue(+5.1)}& 64.5&44.8& 48.0\textsubscript{\blue(+5.2)}&69.5&52.7&43.0&66.8&46.6\\
%
	\bottomrule
\end{tabular}
\end{table}

\subsection{Ablation Studies}

\begin{wraptable}{r}{9cm}
\vskip -0.8in
\caption{Classification performance  for different combinations of LSA (L) and GSA (G) blocks based on the small model.}
\label{tab: abl_global_local}
\smallskip
\centering
\small 
\begin{tabular}{llll}
	\toprule
	Function Type & Params & FLOPs & Top-1\\
	& (M) & (G) & (\%)\\
	\midrule
	(L, L, L) & 8.8& 2.2 & 76.9 \\
	(L, LLG, LLG, G) & 23.5& 2.8 & 81.5 \\
	(L, LG, LG, G) & 24.1& 2.8 &81.7  \\
	(L, L, L, G) & 22.2& 2.9&80.5 \\
	\midrule
	PVT-small (G, G, G, G) \cite{wang2021pyramid} & 24.5& 3.8& 79.8 \\
	\bottomrule
\end{tabular}
\vskip -0.1in

\end{wraptable}

\paragraph{Configurations of LSA and GSA blocks. }
We evaluate different combinations of LSA and GSA based on our small model and present the ablation results in Table~\ref{tab: abl_global_local}. The models with only locally-grouped attention 
fail to 
obtain good performance (76.9\%) because this setting has a limited and small receptive field.
An extra global attention layer in the last stage can improve the classification performance by 3.6\%. Local-Local-Global (\emph{abbr.} LLG) also achieves good performance (81.5\%), but we do not use this design in this work. 

\begin{wraptable}{r}{5.5cm}
\vskip -0.5in
	\centering
\caption{ImageNet classification performance  of different forms of  sub-sampled functions for the global sub-sampled attention (GSA).}
\label{tab: abl_funciton}
\smallskip
\small 
\begin{tabular}{lHHl}
	\toprule
	Function Type & Params(M) & FLOPs(G) & Top-1(\%)\\
	\midrule
	2D Conv. & 23.5& 2.8 & 81.7 \\
	2D Separable Conv. & 23.6& 2.6&81.2 \\
	Average Pooling & 22.2& 2.7& 81.2\\  
	\bottomrule
\end{tabular}

\vskip -0.2in
\end{wraptable}

\paragraph{Sub-sampling functions.}
We further study how the different   sub-sampling   functions affect the performance.  Specifically, we compare the regular strided convolutions, separable convolutions and average pooling based on the `small' model and present the results in Table~\ref{tab: abl_funciton}. The first option performs best and therefore we choose it as our default implementation.

\paragraph{Positional Encodings.}
We replace the relative positional encoding with CPVT for Swin-T and report the detection performance on COCO with RetinaNet and Mask R-CNN in Table~\ref{tab: pos_swin}. The CPVT-based Swin cannot achieve improved performance with both frameworks, which indicates that our performance improvements should be owing to the paradigm of \text{\altour} instead of the positional encodings.
\begin{table}[ht]
 \setlength\tabcolsep{2pt}
	\caption{Object detection performance on the COCO using different positional encoding strategies.}
	\label{tab: pos_swin}
	\centering
\small
\begin{tabular}{r|cclcc|ccccc}
	\toprule
	\multirow{2}{*}{Backbone}  &\multicolumn{5}{c|}{RetinaNet} &\multicolumn{5}{c}{Mask RCNN} \\

	& FLOPs(G) &Param(M)&AP &AP$_{50}$ &AP$_{75}$& FLOPs(G) &Param(M) &AP &AP$_{50}$ &AP$_{75}$  \\
	\midrule
	Swin-T \cite{liu2021swin} &245 &38.5&41.5 & 62.1& 44.2& 264& 47.8& 42.2& 64.6& 46.2\\
	Swin-T+CPVT &245 &38.5&41.3& 62.4&44.1 &263&47.8 & 42.0&64.5&45.9\\
	
	\bottomrule	
\end{tabular}
\end{table}

\section{Conclusion}
In this paper, we have presented  two powerful vision transformer backbones for both image-level classification and a few downstream dense prediction tasks. We dub them as twin transformers: \text{\pcpvt} and \text{\altour}. The former variant explores the applicability of conditional positional encodings \cite{chu2021ConditionalPE} in pyramid vision transformer \cite{wang2021pyramid}, confirming its potential  for improving backbones 
in many vision tasks. 
In the latter variant we revisit current attention design to 
proffer 
a more efficient attention paradigm. We find that interleaving local and global attention can produce impressive results, yet it comes with higher throughputs. \textit{Both transformer models 
set 
a new state of the art in image classification, objection detection and semantic/instance segmentation.}

{
\bibliographystyle{ieee_fullname}

%
\bibliography{CSRef}
	
}

\clearpage

\appendix
\section{Experiment}
\begin{table}[ht]
	\setlength\tabcolsep{0.5pt}
	\caption{Large models' object detection performance on the COCO \texttt{val2017} split using 1$\x$ schedule. } 
	\label{tab: detection-l-1x}
	\centering
	\small 
	\begin{tabular}{rHclcc|Hccl*{7}{c}}
		\toprule
		\multirow{2}{*}{Backbone}  &\multicolumn{5}{c}{RetinaNet 1$\times$} &\multicolumn{5}{c}{Mask R-CNN 1$\times$} \\
		\cmidrule{2-15} 
		&FLOPs(G) &Param(M)&AP &AP$_{50}$ &AP$_{75}$ &FLOPs(G) &Param(M)&AP$^{\rm b}$ &AP$^{\rm m}$ \\
		
		\midrule
		ResNeXt101-64$\times$4d~\cite{xie2017aggregated} & 473 & 95.5& 41.0 & 60.9 & 44.0 &  493& 101.9 &42.8 &38.4 &  &  & & &  \\
		PVT-Large~\cite{wang2021pyramid} & 345& 71.1 & {42.6} & {63.7} & {45.4} &364&81.0&42.9  &39.5   && & & & \\
		\pcpvtlarge \ (ours) &345&71.2&45.1 \textsubscript{\blue(+4.1)}&66.4&48.4 &364&81.2&45.4 \textsubscript{\blue(+2.6)}&41.5\\
		Swin-B \cite{liu2021swin}&477 &98.4&44.7&65.9&47.8& 496& 107.2&45.5&41.3\\
		\altlarge\  (ours) &455 & 110.9&45.7 \textsubscript{\blue(+4.7)}&67.1&49.2&474&119.7&45.9 \textsubscript{\blue(+3.1)}&41.6 \\
		
		\bottomrule	
	\end{tabular}
\end{table}
\section{Algorithm}
\begin{algorithm}[h]
	\caption{PyTorch snippet of LSA.}
	\label{alg:local_group_attention}
	\definecolor{codeblue}{rgb}{0.25,0.5,0.5}
	\lstset{
		backgroundcolor=\color{white},
		basicstyle=\fontsize{7.2pt}{7.2pt}\ttfamily\selectfont,
		columns=fullflexible,
		breaklines=true,
		captionpos=b,
		commentstyle=\fontsize{7.2pt}{7.2pt}\color{codeblue},
		keywordstyle=\fontsize{7.2pt}{7.2pt}\color{blue},
	}
	\begin{lstlisting}[language=python]
class GroupAttention(nn.Module):
    def __init__(self, dim, num_heads=8, qkv_bias=False, qk_scale=None, attn_drop=0., proj_drop=0., k1=7, k2=7):
	super(GroupAttention, self).__init__()
	self.dim = dim
	self.num_heads = num_heads
	head_dim = dim // num_heads
	self.scale = qk_scale or head_dim ** -0.5
	self.qkv = nn.Linear(dim, dim * 3, bias=qkv_bias)
	self.attn_drop = nn.Dropout(attn_drop)
	self.proj = nn.Linear(dim, dim)
	self.proj_drop = nn.Dropout(proj_drop)
	self.k1 = k1
	self.k2 = k2
    
    def forward(self, x, H, W):
        B, N, C = x.shape
        h_group, w_group = H // self.k1, W // self.k2
        total_groups = h_group * w_group
        x = x.reshape(B, h_group, self.k1, w_group, self.k2, C).transpose(2, 3)
        qkv = self.qkv(x).reshape(B, total_groups, -1, 3, self.num_heads, C // self.num_heads).permute(3, 0, 1, 4, 2, 5)
        q, k, v = qkv[0], qkv[1], qkv[2]   
        attn = (q @ k.transpose(-2, -1)) * self.scale
        attn = attn.softmax(dim=-1)
        attn = self.attn_drop(attn)
        attn = (attn @ v).transpose(2, 3).reshape(B, h_group, w_group, self.k1, self.k2, C)
        x = attn.transpose(2, 3).reshape(B, N, C)
        x = self.proj(x)
        x = self.proj_drop(x)
        return x
	\end{lstlisting}
\end{algorithm}
\section{Architecture Setting}
\begin{table}[h]
	\tabcolsep 1.5pt
	\caption{Configuration details of \pcpvt.}
	\label{tab: \pcpvtbase}
	\small 
	\begin{tabular}{*{5}{c|}cc}
		\toprule
		& Output Size & Layer Name &\pcpvtsmall & \pcpvtbase  & \pcpvtlarge &   \\
		\midrule
		\multirow{2}{*}[-2.5ex]{Stage 1} & \multirow{2}{*}[-2.5ex]{\scalebox{1.3}{$\frac{H}{4}\times \frac{W}{4}$}} & Patch Embedding & \multicolumn{4}{c}{$P_1=4$; $C_1=64$} \\
		\cline{3-6}
		& & \tabincell{c}{Transformer\\Encoder with PEG} & 
		$\begin{bmatrix}
			\begin{array}{l}
				R_1=8 \\
				N_1=1 \\
				E_1=8 \\
			\end{array}
		\end{bmatrix} \times 3$ &
		$\begin{bmatrix}
			\begin{array}{l}
				R_1=8 \\
				N_1=1 \\
				E_1=8 \\
			\end{array}
		\end{bmatrix} \times 3$ &
		$\begin{bmatrix}
			\begin{array}{l}
				R_1=8 \\
				N_1=1 \\
				E_1=8 \\
			\end{array}
		\end{bmatrix} \times 3$ \\
		
		\midrule
		\multirow{2}{*}[-2.5ex]{Stage 2} & \multirow{2}{*}[-2.5ex]{\scalebox{1.3}{$\frac{H}{8}\times \frac{W}{8}$}} & Patch Embedding & \multicolumn{4}{c}{$P_2=2$;  $C_2=128$} \\
		\cline{3-6}
		& & \tabincell{c}{Transformer\\Encoder with PEG} &
		$\begin{bmatrix}
			\begin{array}{l}
				R_2=4 \\
				N_2=2 \\
				E_2=8 \\
			\end{array}
		\end{bmatrix} \times 3$ &
		$\begin{bmatrix}
			\begin{array}{l}
				R_2=4 \\
				N_2=2 \\
				E_2=8 \\
			\end{array}
		\end{bmatrix} \times 3$ &
		$\begin{bmatrix}
			\begin{array}{l}
				R_2=4 \\
				N_2=2 \\
				E_2=8 \\
			\end{array}
		\end{bmatrix} \times 8$\\
		\midrule
		\multirow{2}{*}[-2.5ex]{Stage 3} & \multirow{2}{*}[-2.5ex]{\scalebox{1.3}{$\frac{H}{16}\times \frac{W}{16}$}} & Patch Embedding & \multicolumn{4}{c}{$P_3=2$;  $C_3=320$} \\
		\cline{3-6}
		& & \tabincell{c}{Transformer\\Encoder with PEG} &
		$\begin{bmatrix}
			\begin{array}{l}
				R_3=2 \\
				N_3=5 \\
				E_3=4 \\
			\end{array}
		\end{bmatrix} \times 6$ &
		$\begin{bmatrix}
			\begin{array}{l}
				R_3=2 \\
				N_3=5 \\
				E_3=4 \\
			\end{array}
		\end{bmatrix} \times 18$ &
		$\begin{bmatrix}
			\begin{array}{l}
				R_3=2 \\
				N_3=5 \\
				E_3=4 \\
			\end{array}
		\end{bmatrix} \times 27$\\
		\midrule
		\multirow{2}{*}[-2.5ex]{Stage 4} &  \multirow{2}{*}[-2.5ex]{\scalebox{1.3}{$\frac{H}{32}\times \frac{W}{32}$}} & Patch Embedding & \multicolumn{4}{c}{$P_4=2$; $C_4\!=\!512$} \\
		\cline{3-6}
		& & \tabincell{c}{Transformer\\Encoder with PEG} 
		&
		$\begin{bmatrix}
			\begin{array}{l}
				R_4=1 \\
				N_4=8 \\
				E_4=4 \\
			\end{array}
		\end{bmatrix} \times 3$ & $\begin{bmatrix}
			\begin{array}{l}
				R_4=1 \\
				N_4=8 \\
				E_4=4 \\
			\end{array}
		\end{bmatrix} \times 3$ & $\begin{bmatrix}
			\begin{array}{l}
				R_4=1 \\
				N_4=8 \\
				E_4=4 \\
			\end{array}
		\end{bmatrix} \times 3$\\
		\bottomrule
	\end{tabular}
\end{table}

\begin{table}[t]
	\tabcolsep 1pt
	\caption{
		Configuration details of \altour.}
	\label{tab: \altour}
	\small 
	\centering 
	\begin{tabular}{*{5}{c|}cc}
		\toprule
		& Output Size & Layer Name &\altsmall & \altbase  & \altlarge &   \\
		\midrule
		\multirow{2}{*}[-2.5ex]{Stage 1} & \multirow{2}{*}[-2.5ex]{\scalebox{1.3}{$\frac{H}{4}\times \frac{W}{4}$}} & Patch Embedding &$P_1=4$;  $C_1=64$ &$P_1=4$; $C_1=96$ & $P_1=4$; $C_1=128$\\
		\cline{3-6}
		& & \tabincell{c}{Transformer\\Encoder w/ PEG} & 
		$\begin{bmatrix}
			\begin{array}{l}
				LSA \\
				GSA \\
				
			\end{array}
		\end{bmatrix} \times 1$ &
		$\begin{bmatrix}
			\begin{array}{l}
				LSA \\
				GSA \\
			\end{array}
		\end{bmatrix} \times 1$ &
		$\begin{bmatrix}
			\begin{array}{l}
				LSA \\
				GSA \\
			\end{array}
		\end{bmatrix} \times 1$ \\
		
		\midrule
		\multirow{2}{*}[-2.5ex]{Stage 2} & \multirow{2}{*}[-2.5ex]{\scalebox{1.3}{$\frac{H}{8}\times \frac{W}{8}$}} & Patch Embedding & $P_2=2$;  $C_2=128$&$P_2=2$;  $C_2=192$&$P_2=2$;  $C_2=256$  \\
		\cline{3-6}
		& & \tabincell{c}{Transformer\\Encoder w/ PEG} &
		$\begin{bmatrix}
			\begin{array}{l}
				LSA\\
				GSA\\
			\end{array}
		\end{bmatrix} \times 1$ &
		$\begin{bmatrix}
			\begin{array}{l}
				LSA\\
				GSA\\
			\end{array}
		\end{bmatrix} \times 1$ &
		$\begin{bmatrix}
			\begin{array}{l}
				LSA\\
				GSA\\
			\end{array}
		\end{bmatrix} \times 1$\\
		\midrule
		\multirow{2}{*}[-2.5ex]{Stage 3} & \multirow{2}{*}[-2.5ex]{\scalebox{1.3}{$\frac{H}{16}\times \frac{W}{16}$}} & Patch Embedding & $P_3=2$;  $C_3=256$& $P_3=2$;  $C_3=384$& $P_3=2$;  $C_3=512$\\
		\cline{3-6}
		& & \tabincell{c}{Transformer\\Encoder w/ PEG} &
		$\begin{bmatrix}
			\begin{array}{l}
				LSA\\
				GSA \\
			\end{array}
		\end{bmatrix} \times 5$ &
		$\begin{bmatrix}
			\begin{array}{l}
				LSA\\
				GSA \\
			\end{array}
		\end{bmatrix} \times 9$ &
		$\begin{bmatrix}
			\begin{array}{l}
				LSA\\
				GSA \\
			\end{array}
		\end{bmatrix} \times 9$\\
		\midrule
		\multirow{2}{*}[-2.5ex]{Stage 4} &  \multirow{2}{*}[-2.5ex]{\scalebox{1.3}{$\frac{H}{32}\times \frac{W}{32}$}} & Patch Embedding &$P_4=2$; $C_4\!=\!512$ & $P_4=2$; $C_4\!=\!768$&$P_4=2$; $C_4\!=\!1024$\\
		\cline{3-6}
		& & \tabincell{c}{Transformer\\Encoder w/ PEG} 
		&
		$\begin{bmatrix}
			\begin{array}{l}
				GSA\\
			\end{array}
		\end{bmatrix} \times 4$ & $\begin{bmatrix}
			\begin{array}{l}
				GSA\\
			\end{array}
		\end{bmatrix} \times 2$ & $\begin{bmatrix}
			\begin{array}{l}
				GSA\\
			\end{array}
		\end{bmatrix} \times 2$\\
		\bottomrule
	\end{tabular}
\end{table}


\end{document}